\documentclass{article}

\usepackage[preprint]{neurips_2019}

\usepackage{algorithm}
\usepackage{algorithmic}
\usepackage{amsmath}
\usepackage{amsthm}
\usepackage{amssymb}
\usepackage{float}
\usepackage{hyperref}

\usepackage[english]{babel}
\usepackage{graphicx}
\usepackage{tikz,graphics,color,fullpage,float,epsf,caption,subcaption}

\usepackage[utf8]{inputenc} 
\usepackage[T1]{fontenc}    
\usepackage{hyperref}       
\usepackage{url}            
\usepackage{booktabs}       
\usepackage{amsfonts}       
\usepackage{nicefrac}       
\usepackage{microtype}      

\def\E{\mathbb E}

\newtheorem{theorem}{Theorem}[section]
\newtheorem{example}[theorem]{Example}
\newtheorem{remark}[theorem]{Remark}

\title{Active Learning with Importance Sampling}

\author{
  Muni Sreenivas Pydi \\
  Department of Electrical \& Computer Engineering\\
  University of Wisconsin-Madison\\
  \texttt{pydi@wisc.edu} \\  
  \And
  Vishnu Suresh Lokhande \\
  Department of Computer Science\\
  University of Wisconsin-Madison\\
  \texttt{lokhande@cs.wisc.edu} \\  
}

\begin{document}

\maketitle

\begin{abstract}
We consider an active learning setting where the algorithm has access to a large pool of unlabeled data and a small pool of labeled data. In each iteration, the algorithm chooses few unlabeled data points and obtains their labels from an oracle. In this paper, we consider a probabilistic querying procedure to choose the points to be labeled. 
We propose an algorithm for Active Learning with Importance Sampling (ALIS), and derive upper bounds on the true loss incurred by the algorithm for any arbitrary probabilistic sampling procedure. Further, we propose an optimal sampling distribution that minimizes the upper bound on the true loss.
\end{abstract}

\section{Introduction}

Active learning is an important machine learning paradigm with a rich class of problems and mature literature \citep{prince2004does,settles2012active,hanneke2014theory}. Oftentimes, users have access to a large pool of unlabeled data and an oracle that can provide a label to a data point that is queried. Querying the oracle for the label comes at a cost, computational and/or monetary. Hence, a key objective for the algorithm is to ``wisely'' choose the set of points from the unlabelled pool that can provide better generalization. In this paper, we propose a probabilistic querying procedure to choose the points to be labeled by the oracle motivated from importance sampling literature \citep{tokdar2010importance}. Importance sampling  is a popular statistical technique widely used for fast convergence in Monte Carlo based methods \citep{doucet2001introduction} and stochastic optimization \citep{zhao2015stochastic}. 

The \textit{main contributions }of this paper are as follows. (a) We propose an importance sampling based algorithm for active learning, which we call Active Learning with Importance Sampling (ALIS). (b) We derive a high probability upper bound on the true loss and design the ALIS algorithm to directly minimize the bound. (c) We determine an optimal sampling probability distribution for the algorithm. (d) We demonstrate that the optimal sampling distribution gives a tighter bound on the true loss compared to the baseline uniform sampling procedure.


\section{Setting}

We consider a binary classification problem where there is a large pool of $n$ $i.i.d.$ data points $\{(x_i, y_i)\}_{i \in [n]} \sim \mathcal{Z}$ drawn from an underlying probability distribution $\mathcal{Z} = \mathcal{X} \times \mathcal{Y}$, out of which, the labels are available only for a small pool of data points $S^0 = \{s^0(j) \in [n]\}_{j\in [m]}$.

At the start, an active algorithm $\mathcal A^0$ has access to all the data $T = \{x_i\}_{i\in [n]}$ in the large pool and the labels $\{y_{s^0(j)}\}_{j\in [m]}$ for the data in the small pool $S^0$. For the next iteration, a sampling procedure is used to select a small subset, $S^1 = \{s^1(j) \in [n]\}_{j\in [m_1]}$ of size $m_1$, for which the labels are requested from the oracle. The algorithm $\mathcal A^1$ now has access to the labels for the data points in $S^0 \cup S^1$ and subsequently requests labels for yet another small subset $S^2$. 
Let $m_t$ denote the number of data points to be queried by the oracle in iteration $t$. Let $n_t$ denote the number of unlabeled points after $t$ iterations.
At the end of $t$ iterations, algorithm $\mathcal A^t$ has access to labels for the data points in $S^0 \cup \ldots \cup S^t$, and samples $m_t$ data points from the remaining 
$n_t= n - \sum_{s=1}^{t-1} m_s$ unlabeled data points.
Let $l(z_i, \mathcal A^t)$ be the loss evaluated on the algorithm $A^t$ for the data point $z_i := (x_i, y_i)$.  Let $U^t = \{u^t(j) \in [n]\}_{j \in [n_t]}$ denote the set of unlabeled points after $t$ iterations, and $V_t\subseteq U_t$ denote the set of points picked by algorithm $\mathcal A^t$ for querying.

\section{Problem Formulation and Approach}\label{sec_approach}

We consider a probabilistic sampling procedure to select the data points for labeling by the oracle. Let $\{p_{u^t(j)}^t\}_{i\in [n_t]}$ denote the probabilities of querying the labels for the unlabeled points in $U^t$ at iteration $t$. For convenience, we write $p^t$ to denote this probability distribution.
Let $\{Q_i^t\}_{i\in [n_t]}$ denote a set of Bernoulli random variables such that $\mathbb P(Q_i^t =1) = p_i^t$. A label for  $x_i$ is requested from the oracle at time $t$, only when $Q_i^t =1$.  
Our goal is to derive the optimal querying probability distribution $p^t$ that minimizes the true loss $\E_{\mathcal Z}[l(z,\mathcal A^t)]$ of the algorithm $\mathcal A^t$ at every iteration $t$. We begin by writing an upper bound for the true loss as follows. This decomposition of the true loss  is inspired from \cite{iclr18}, albeit generalized to account for non-uniform sampling using a weighted loss across the queried data points.

\begin{align}\label{eq_coreset}
	\E_{\mathcal Z}[l(z;\mathcal A^t)] 
	&\le 
	\underbrace{\left| \E_{\mathcal Z}[l(z,\mathcal A^t))] - \frac{1}{n_t} \sum_{j \in [n_t]} l(z_{u^t(j)},\mathcal A^t) \right|}_{(A)}
    + 
    \underbrace{\frac{1}{n_t} \sum_{j \in [n_t]} \frac{Q_{u^t(j)}^t}{p_{u^t(j)}^t} l(z_{u^t(j)},\mathcal A^t)}_{(B)}\\    \nonumber
    &+ 
    \underbrace{\left| \frac{1}{n_t} \sum_{j \in [n_t]} l(z_{u^t(j)},\mathcal A^t) - \frac{1}{n_t} \sum_{j \in [n_t]} \frac{Q_{{u^t(j)}}^t}{p_{u^t(j)}^t} l(z_{u^t(j)},\mathcal A^t) \right|}_{(C)}.
\end{align}

In equation~\eqref{eq_coreset},  $(A)$ corresponds to the generalization error of the algorithm $\mathcal A^t$ on all the $n_t$ unlabeled data points.  $(B)$ corresponds to a weighted average loss for $\mathcal A^t$ over the data points that will be queried at the end of $t$ iterations, i.e. the points for which $Q_i^t=1$.  $(C)$ is the absolute difference between the average loss over the all the $n_t$ unlabeled data points and the weighted average loss for the queried data points.

In the term $(B)$, the loss at $z_{u^t(j)}$ is scaled by $1/{p_{u^t(j)}^t}$ if $Q_{{u^t(j)}}^t = 1$, i.e. if it is chosen for sampling at iteration $t$.
Since $\mathbb E[Q_i^t] = p_i^t$, we have
\begin{align*}
\mathbb E_{Q^t}\left[\frac{1}{n_t} \sum_{j \in [n_t]} \frac{Q_{{u^t(j)}}^t}{p_{u^t(j)}^t} l(z_{u^t(j)},\mathcal A^t) \right] =
\frac{1}{n_t} \sum_{j \in [n_t]} l(z_{u^t(j)},\mathcal A^t)
\end{align*}
where $\mathbb E_{Q^t}[.]$ denotes the expectation with respect to the random variables $\{Q_{u_t(j)}^t\}_{j\in [n_t]}$. Hence, the weighted average loss in $(B)$ is an {\em unbiased estimate} of the average loss over all the unlabeled data points in $U^t$. 
Moreover, since the data points are $i.i.d.$, $(C)$ can be expressed as the sum of $n_t$ zero mean $i.i.d.$ random variables denoted by $H^t_j = l(z_{u^t(j)},\mathcal A^t) -  \frac{Q_{{u^t(j)}}^t}{p_{u^t(j)}^t} l(z_{u^t(j)},\mathcal A^t)$, whose variance is given by,
\begin{align}\label{eq_variance}
 \mathop{Var}(H_j^t) 
 = \frac{l(z_{u^t(j)},\mathcal A^t)^2}{\left(p_{u^t(j)}^t\right)^2} \mathop{Var}\left( Q_{{u^t(j)}}^t \right)
 = l(z_{u^t(j)},\mathcal A^t)^2
 \left(\frac{1}{p_{u^t(j)}^t}-1\right).
 \end{align}

The choice of the sampling distribution $p^t$ does not affect $(A)$, but affects $(B)$ and $(C)$. Term $(C)$ is zero in expectation over the random variables $\{Q_j^t\}_{j\in [n_t]}$, and its variance directly depends on the 
sampling distribution $p^t$. Hence, by choosing a sampling distribution that minimizes the variance of term $(C)$, we can get a tight bound on the true loss. 
Given a sufficiently expressive hypothesis space and an algorithm $A^t$ that is designed to account for weighted loss,  $(B)$ can be driven to zero via weighted empirical risk minimization. Hence, choosing the optimal $p^t$ (in the sense of minimizing the variance in \eqref{eq_variance}) minimizes the true loss, $\E_{\mathcal Z}[l(z,\mathcal A^t)]$.

However, the variance of $(C)$ depends on the true loss $l(z_{u^t(j)},\mathcal A^t)$ on unseen data points. Hence, minimizing the variance in \eqref{eq_variance} directly w.r.t $p^t$ is not possible. 
We get around this problem using an approach suggested by \cite{wang2015querying} wherein a ``pseudo-label'' is used to get an upper bound on the actual loss, $l(z_{u^t(j)},\mathcal A^t)$. Suppose $f_{\mathcal A^t}$ be the classifier learnt by $\mathcal A^t$. 
We can upper bound $l(z_{u^t(j)},\mathcal A^t)$ by $l(\hat{z}_{u^t(j)},\mathcal A^t)$ 
where $\hat{z}_{u^t(j)} = (x_{u^t(j)}, \hat{y}_{u^t(j)})$ 
and $\hat{y}_{u^t(j)} = -\mathop{sign}(f(x_{u^t(j)}))$.
We call $l(\hat{z}_{u^t(j)},\mathcal A^t)$ the ``pseudo-loss'' at the unlabeled data point $x_{u^t(j)}$. The pseudo label is designed so that the algorithm $\mathcal A^t$ always suffers a worst loss on the pseudo label than the true label. Hence,  
$l(z_{u^t(j)},\mathcal A^t)\leq l(\hat{z}_{u^t(j)},\mathcal A^t)$

\section{Related Work}
The closest work to our setting is \cite{beygelzimer2009importance} which considers an importance weighted sampling procedure for active learning. However, the proposed algorithm does not have the freedom to choose the data point to be labeled. Instead, the algorithm receives unlabeled data points in an online fashion and makes a choice whether or not to query the oracle for their labels. 
In our setting, an unlabeled pool of data is available from the start, and the choice is over which data points to be queried for labels. Having the entire unlabeled pool at hand allows for us to tune the sampling distribution based on additional information about the unlabeled points, for example, the pseudo-loss evaluated on the unlabeled points. 
Moreover, the weighted sampling procedure proposed in \cite{beygelzimer2009importance} is based on a version space approach over a finite hypothesis class. Our setting does not have such restrictions.
In \cite{iclr18}, the authors use a very similar true error decomposition as in \eqref{eq_coreset}, but use a deterministic procedure based on coresets to choose the points to be labeled. Our approach is probabilistic, and computationally far less intensive.

\section{Algorithm}

In this section, we present an algorithmic framework for Active Learning with Importance Sampling (ALIS). For a specific choice of sampling distribution $\{p^t\}_{t=1}^T$, the algorithm proceeds as follows.

\begin{algorithm}
\label{alg1}
\caption{Active Learning with Importance Sampling (ALIS)}
\begin{algorithmic}\label{alg1}
\REQUIRE $S^0$ (labeled dataset)
\REQUIRE $U^0$ (Unlabeled dataset)
\FOR{ $t= 1, 2, \hdots, T$}
\STATE Compute pseudo-labels $\hat y = -\mathop{sign}\left(f_{\mathcal A^t}(x)\right)$ for data points in $U^t$ using algorithm $\mathcal A^t$
\STATE Compute $p^t$ as in \eqref{eq_prob} using loss computed on pseudo-labels.
\STATE Sample data points from $U^t$ according to the  distribution $p^t$ to form a new set of points, $V^t$.
\STATE Receive true labels for data points in $V^t$ from the oracle.
\STATE Update $S^t$: $S^{t+1} \leftarrow S^t \cup V^t$
\STATE Update $U^t$: $U^{t+1} \leftarrow U^t \backslash V^t$
\STATE Retrain algorithm  on $V^t$ to minimize the weighted error $\sum_{x_k \in V^t} \frac{1}{p_k^t}l(z_k, \mathcal A^{t+1})$ to get $A^{t+1}$.
\ENDFOR
\RETURN Algorithm $\mathcal A^{T+1}$
\end{algorithmic}
\label{algo:framework}
\end{algorithm}

\section{Analysis}

In this section, we derive an upper bound on the true loss (i.e. expected loss over the entire domain of the data generating distribution $\mathcal Z$) for Algorithm \ref{alg1}, that holds with high probability over the random sampling using $p^t$.

\begin{theorem}\label{thm_genBound}
Define $M_p^t := \sum_{j\in [n_t]} \frac{l(\hat{z}_{u^t(j)} ,\mathcal A^t)}{p_{u^t(j)}^t}$. 
Let  $c_\delta>0$ be a constant that depends on $\delta$.
Using Algorithm \ref{alg1}, the true error at the end of $t$ iterations is  bounded as follows, with probability at least $1 - \delta$.
\begin{align}\label{eq_bound}
\E_{\mathcal Z}[l(x,y;\mathcal A^t)] 
&\le 
\left| 
\E_{\mathcal Z}[l(x,y;\mathcal A^t))] - \frac{1}{n_t} \sum_{j \in [n_t]} l(\hat{z}_{u^t(j)} ,\mathcal A^t) \right| 
+ c_\delta \frac{M_p^t}{n_t}.
\end{align}
\end{theorem}
\begin{proof}
See Appendix~\ref{app_proofs}.
\end{proof} 

\begin{remark}
The bound in \eqref{eq_bound} is valid for any arbitrary sampling distribution $p^t$, via $M_p^t$. While the first term in \eqref{eq_bound} is the generalization error which is independent of $p^t$, the second term, $M_p^t/n_t$, is the weighted average pseudo-loss on the unlabeled dataset at iteration $t$ and can be computed in practice using pseudo-labels.
\end{remark}

It is clear from \eqref{eq_bound} that choosing $p^t$ so as to minimize $M_p^t$ will result in the tightest bound for the expected loss of Algorithm~\ref{alg1}. In the next theorem, we present the optimal sampling probability distribution $(p^*)^t$ that minimizes $M^t$. 

\begin{theorem}\label{thm_1}
The optimal distribution $(p^*)^t$ for minimizing $M_p^t = \sum_{j\in [n_t]} \frac{l(\hat{z}_{u^t(j)} ,\mathcal A^t)}{p_{u^t(j)}^t}$ is given by
\begin{align}\label{eq_prob}
(p^*)_{u^t(j)}^t = \frac{l(\hat{z}_{u^t(j)},\mathcal A^t)^{1/2}}{\sum_{j\in [n_t]} l(\hat{z}_{u^t(j)},\mathcal A^t)^{1/2}}.
\end{align}
\end{theorem}
\begin{proof}
See Appendix~\ref{app_proofs}.
\end{proof}

\begin{example}
From ~\eqref{eq_prob}, we see that the optimal probability of querying a data point for its label is proportional to the square root of pseudo-loss evaluated  at that point. For example, consider the squared error loss function, $l(z, \mathcal A) = \left( y - f_{\mathcal A}(x)\right)^2$. The pseudo loss at $z$ is given by 
$l(\hat{z}, \mathcal A) = \left( \hat{y} - f_{\mathcal A}(x)\right)^2 = \left( -\mathop{sign}(f_{\mathcal A}(x)) - f_{\mathcal A}(x)\right)^2 = \left( 1 + |f_{\mathcal A}(x)|\right)^2$. Hence, in this case  the optimal querying probability in \eqref{eq_prob} reduces to $p_{u^t(j)}^t = \frac{1+|f_{\mathcal A^t}(x_{u^t(j)})|}{\sum_{j\in [n_t]}  1+|f_{\mathcal A^t}(x_{u^t(j)})| }$.
\end{example}

\begin{remark}
To see that sampling using the optimal distribution $(p^*)^t$ indeed gives a tighter bound on true loss compared to uniform sampling, let us compare $M_{p^*}^t$ with $M_q^t$, where $q^t$ is a uniform distribution over unlabeled points at iteration $t$, i.e. $q^t_{u^t(j)} = 1/n_t$. Then we have,
\begin{align*}
    M_{p^*}^t 
    = 
    \sum_{j\in [n_t]} \frac{l(\hat{z}_{u^t(j)} ,\mathcal A^t)}{(p^*)_{u^t(j)}^t}
    &= 
    \sum_{j\in [n_t]} \frac{l(\hat{z}_{u^t(j)} ,\mathcal A^t)}{l(\hat{z}_{u^t(j)} ,\mathcal A^t)^{1/2}}  
    \sum_{j\in [n_t]} l(\hat{z}_{u^t(j)} ,\mathcal A^t)^{1/2}\\
    &=
    \left( \sum_{j\in [n_t]} l(\hat{z}_{u^t(j)} ,\mathcal A^t)^{1/2} \right)^2\\
    &\leq 
    \left( \sum_{j\in [n_t]} 1\right)
    \left( \sum_{j\in [n_t]} l(\hat{z}_{u^t(j)} ,\mathcal A^t)\right)
    =  \sum_{j\in [n_t]}
    \frac{l(\hat{z}_{u^t(j)} ,\mathcal A^t)}{1/n_t}
    = M_q^t.
\end{align*}
Here, the inequality follows from Cauchy-Schwartz. Hence, we have shown that $M_{p^*}^t \leq M_q^t$.
\end{remark}

\section{Conclusion}
In this paper, we propose an algorithm for Active Learning with Importance Sampling (ALIS) which uses a non-uniform probability distribution to select points to be queried from the oracle. 
We derive an upper bound on the true loss incurred by an ALIS algorithm for any arbitrary sampling procedure.
Our bound depends on the generalization error, and the weighted average pseudo-loss computed on the data that is yet to be seen.
We then propose a sampling procedure that is optimal for minimizing the upper bound on the true loss and show that sampling using the optimal probability distribution gives a tighter bound on the true loss compared to uniform sampling.

\small

\bibliographystyle{plainnat}
\bibliography{ms}

\appendix

\section{Proofs}\label{app_proofs}

\begin{proof}[Proof of Theorem~\ref{thm_genBound}]
Define 
$H^t_j = l({z}_{u^t(j)} ,\mathcal A^t) -  \frac{Q_{{u^t(j)}}^t}{p_{u^t(j)}^t} l({z}_{u^t(j)} ,\mathcal A^t)$.
For any choice of $p^t$, we get $\E[H_j^t]=0$ for all $j\in [n_t]$. 
In addition, $H_j^t$'s are bounded as follows.
\begin{align*}
    |H_j^t|
    = 
    l({z}_{u^t(j)} ,\mathcal A^t)
    \left\lvert 1-\frac{Q_{{u^t(j)}}^t}{p_{u^t(j)}^t}
    \right\rvert    
    \leq
    \frac{l(\hat{z}_{u^t(j)} ,\mathcal A^t)}{p_{u^t(j)}^t}
    \left\lvert p_{u^t(j)}^t - Q_{{u^t(j)}}^t
    \right\rvert    
    \leq
    \frac{l(\hat{z}_{u^t(j)} ,\mathcal A^t)}{p_{u^t(j)}^t}
    \leq M_p^t.
\end{align*}

Let $S^t := \sum_{j\in [n_t]} \E\left[ \left( H_j^t\right)^2\right]$. Using \eqref{eq_variance}, we get the following bound on $S^t$.
\begin{align*}
    S^t = 
    \sum_{j\in [n_t]} 
    \mathop{Var}\left[ \left( H_j^t\right)^2\right]
    &= \sum_{j\in [n_t]} 
    l(z_{u^t(j)},\mathcal A^t)^2
    \left(\frac{1}{p_{u^t(j)}^t}-1\right)\\
    &\leq \sum_{j\in [n_t]} 
    \frac{l(\hat{z}_{u^t(j)},\mathcal A^t)^2}{p_{u^t(j)}^t}\\
    &\stackrel{(*)}{\leq} 
    \max_{j\in [n_t]} l(\hat{z}_{u^t(j)} ,\mathcal A^t)
    \sum_{j\in [n_t]} 
    \frac{l(\hat{z}_{u^t(j)},\mathcal A^t)}{p_{u^t(j)}^t}\\
    &\leq (M_p^t)^2.
\end{align*}
Here, $(*)$ follows from Holder's inequality. 

Since the data points for querying are chosen independently (with replacement), the random variables $Q_i^t$'s are independent and so are $H_j^t$'s.
Since $H_j^t$'s are independent bounded random variables with finite variance, we can use Bernstein's inequality to get 
$\mathbb{P}\left( \sum_{j\in [n_t]} H^t_j > \epsilon \right)< e^{-\epsilon^2/(2S^t + \frac{2}{3}M_p^t \epsilon)}$. 
Making an $(\epsilon, \delta)$ switch, with probability at least $1-\delta$, we have the following bound on the term $(C)$ in \eqref{eq_coreset}.
\begin{align*}
    \left| \frac{1}{n_t} \sum_{j \in [n_t]} l(z_{u^t(j)},\mathcal A^t) - \frac{1}{n_t} \sum_{j \in [n_t]} \frac{Q_{{u^t(j)}}^t}{p_{u^t(j)}^t} l(z_{u^t(j)},\mathcal A^t) \right|
    &=
    \frac{1}{n_t} \sum_{j \in [n_t]} H_j^t\\
    &\leq
    \frac{M_p^t}{3n_t}\log\frac{1}{\delta}
    \left( 
    1+\sqrt{1+ \frac{18S^t}{(M_p^t)^2\log\frac{1}{\delta}}}
    \right)\\
    &\leq 
    \frac{M_p^t}{3n_t}\log\frac{1}{\delta}
    \left( 
    1+\sqrt{1+ \frac{18}{\log\frac{1}{\delta}}}
    \right).
\end{align*}

Now we bound the term $(B)$ in \eqref{eq_coreset} as follows.
\begin{align}
\frac{1}{n_t} \sum_{j \in [n_t]} \frac{Q_{u^t(j)}^t}{p_{u^t(j)}^t} l({z}_{u^t(j)} ,\mathcal A^t)
\leq 
\frac{1}{n_t} \sum_{j \in [n_t]} \frac{Q_{u^t(j)}^t}{p_{u^t(j)}^t} l(\hat{z}_{u^t(j)} ,\mathcal A^t)
\leq 
\frac{1}{n_t} \sum_{j \in [n_t]} \frac{l(\hat{z}_{u^t(j)} ,\mathcal A^t)}{p_{u^t(j)}^t} 
=  \frac{M_p^t}{n_t}.
\end{align}
Combining the bounds on $(B)$ and $(C)$ we get the desired bound in \eqref{eq_bound}.
\end{proof}

\begin{proof}[Proof of Theorem~\ref{thm_1}]
Since $p^t$ is a probability distribution over $U^t$, we have $\sum_{j\in [n_t]}p^t_{u^t(j)} = 1$. Using $\lambda$ as the Lagrangian variable for this equality constraint, we get the following Lagrangian for $p^t_{u^t(j)}$.
\begin{align*}
L\left( p^t, \lambda\right)
= \sum_{j\in [n_t]}\frac{l(\hat{z}_{u^t(j)},\mathcal A^t)}{p_{u^t(j)}^t} 
+ \lambda\left(\sum_{j\in [n_t]}p^t_{u^t(j)} - 1  \right).
\end{align*}
Equating $\frac{\partial L}{\partial p^t_{u^t(j)}}$ to zero, we get 
\begin{align*}
-\frac{l(\hat{z}_{u^t(j)},\mathcal A^t)}{{p_{u^t(j)}^t}}
+\lambda = 0\ \ 
\Rightarrow p_{u^t(j)}^t 
= \frac{l(\hat{z}_{u^t(j)},\mathcal A^t)^{1/2}}{\sqrt{\lambda}}
\end{align*}
Applying the equality constraint, we get \eqref{eq_prob}.
\end{proof}

\end{document}